\documentclass[letterpaper, 10 pt, conference]{ieeeconf}  % Comment this line out if you need a4paper

\IEEEoverridecommandlockouts                              % This command is only needed if 
\usepackage{graphicx}
\usepackage{amsmath}
\usepackage{amssymb}
\usepackage{booktabs}
\usepackage{float}
\usepackage{color}
\usepackage{tabularray}
\usepackage{caption}
\usepackage{xcolor}
\usepackage{svg}
\usepackage{booktabs}
\usepackage{multirow}
\usepackage{cite}
\usepackage[breaklinks,hidelinks]{hyperref}
\usepackage[nolist,nohyperlinks]{acronym}
\usepackage{siunitx}
\usepackage{cleveref}

\acrodef{rl}[RL]{Reinforcement Learning}
\acrodef{mlp}[MLP]{Multi-Layer Perceptron}
\acrodef{lstm}[LSTM]{Long Short-Term Memory}
\acrodef{fov}[FOV]{Field-of-View}
\acrodef{ppo}[PPO]{Proximal Policy Optimization}
\acrodef{appo}[APPO]{Asynchronous Proximal Policy Optimization}
\acrodef{rnn}[RNN]{Recurrent Neural Network}
\acrodef{gru}[GRU]{Gated Recurrent Unit}
\acrodef{pomdp}[POMDP]{Partially Observable Markov Decision Process}
\acrodef{cmwae}[CMWAE]{Cross-Modal Wasserstein Autoencoder}
\acrodef{mmd}[MMD]{Maximum Mean Discrepancy}
\acrodef{vae}[VAE]{Variational Autoencoder}
\acrodef{wae}[WAE]{Wasserstein Autoencoder}
\acrodef{imu}[IMU]{Inertial Measurement Unit}
\acrodef{mse}[MSE]{Mean Squared Error}
\acrodef{ssim}[SSIM]{Structural Similarity Index Measure}
\acrodef{dce}[DCE]{Deep Collision Encoder}
\acrodef{fc}[FC]{Fully Connected}

\title{\LARGE \bf
Cross-Modal Reinforcement Learning for Navigation \\ with Degraded Depth Measurements
}

\author{Omkar Sawant$^*$, Luca Zanatta, Grzegorz Malczyk, and Kostas Alexis% <-this % stops a space
\thanks{This work was supported by Horizon Europe under Grant EC 101120732 and the Research Council of Norway under Award NO-338694. The authors are with the Department of Engineering Cybernetics, Norwegian University of Science and Technology (NTNU), Norway.} 
\thanks{$^*$Corresponding author. Email: \tt\small omkar.sawant@ntnu.no}
}

\begin{document}

\maketitle
\thispagestyle{empty}
\pagestyle{empty}

%%%%%%%%%%%%%%%%%%%%%%%%%%%%%%%%%%%%%%%%%%%%%%%%%%%%%%%%%%%%%%%%%%%%%%%%%%%%%%%%
\begin{abstract}

This paper presents a cross-modal learning framework that exploits complementary information from depth and grayscale images for robust navigation. We introduce a Cross-Modal Wasserstein Autoencoder that learns shared latent representations by enforcing cross-modal consistency, enabling the system to infer depth-relevant features from grayscale observations when depth measurements are corrupted. The learned representations are integrated with a Reinforcement Learning-based policy for collision-free navigation in unstructured environments when depth sensors experience degradation due to adverse conditions such as poor lighting or reflective surfaces. Simulation and real-world experiments demonstrate that our approach maintains robust performance under significant depth degradation and successfully transfers to real environments.
% We propose a deep navigation policy that combines a Cross-modal Wasserstein Autoencoder with Asynchronous Proximal Policy Optimization based RL framework to enable robust control in the case of sensor degradation. Our framework encodes multimodal inputs — grayscale images and depth maps —into a shared latent representation that is resilient to modality dropouts. The CMWAE is trained with a Wasserstein distance objective which yields a compact and generative latent space that captures important scene geometry. This latent representation, along with robot state information, is used to train a reinforcement learning policy capable of reliable navigation even when sensory inputs are corrupted with a considerable amount of invalid pixels. A set of experimental studies demonstrate the robustness and effectiveness of the proposed approach in both simulated and real-world environments.

\end{abstract}
\vspace{1mm}
%%%%%%%%%%%%%%%%%%%%%%%%%%%%%%%%%%%%%%%%%%%%%%%%%%%%%%%%%%%%%%%%%%%%%%%%%%%%%%%%

\section{INTRODUCTION}

% Robots operating in the real world must be capable of making safe decisions despite degradation or loss of sensory inputs. This is especially true for aerial robots navigating complex, cluttered and GPS denied environments, such as subterranean tunnels, dense forests and disaster-stricken building \cite{tranzatto2022cerberus, delmerico2019rescue}. Conventional navigation systems, relying on accurate maps often fail in such cases due sensor malfunction. Furthermore, the separation of motion planning from control further limits robustness. Recently, there has been an emergence of navigation policies which don't rely on maps of the environment to achieve collision free navigation, such as \cite{loquercio2018dronet}, to tackle these issues. The works along this line do not leverage any cross modality techniques to compensate for sensor malfunction. 

Autonomous navigation in unstructured environments remains a fundamental challenge in robotics, particularly when robots must operate under degraded or incomplete sensory information. Aerial robots deployed in GPS-denied settings, such as subterranean tunnels or disaster zones, face severe limitations in depth sensing due to poor lighting conditions, reflective surfaces, or sensor failures \cite{tranzatto2022cerberus, delmerico2019rescue}. These challenges render traditional map-based navigation pipelines unreliable, as they depend critically on consistent and accurate depth measurements for obstacle detection and path planning. Classical approaches to autonomous navigation typically decouple perception, planning, and control into separate modules \cite{siciliano2016robotics}. While this modularity offers certain advantages, it also introduces fragility: failures in any single module can cascade through the system, leading to unsafe behavior. The reliance on explicit map representations further compounds this vulnerability, as map quality directly depends on sensor reliability~\cite{presentfutureslamextreme}.
%When depth sensors fail or produce corrupted measurements, the entire navigation pipeline breaks down. -- this is wrong. 

\begin{figure}
    \centering
    \includegraphics[width=\columnwidth]{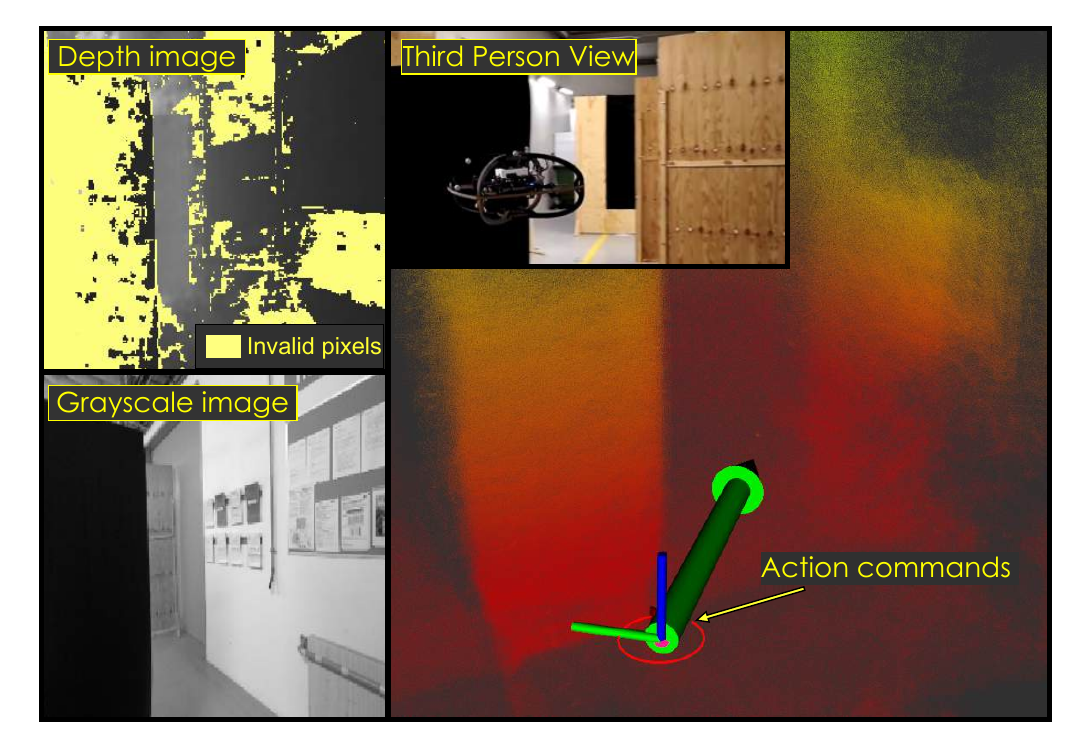}
    \vspace{-4ex}
    \caption{Real-world navigation under depth sensor degradation with preserved grayscale image.}
    \vspace{-4ex}
    \label{fig:title_image}
\end{figure}
Recent advances in learning-based navigation have demonstrated the potential of end-to-end policies that directly map sensory observations to control commands without requiring explicit maps \cite{loquercio2018dronet, kulkarni2024rl,loquercio2021learning}. These approaches offer an extra layer of robustness by learning implicit representations of navigable space. However, when depth information becomes unavailable or corrupted, these systems have no fallback strategy, severely limiting their applicability in real-world scenarios where sensor failures are common \cite{zhang2023perception}.

To address this gap, we propose a cross-modal learning framework that exploits complementary information from multiple sensor modalities. Our key insight is that even when depth measurements degrade, other modalities, particularly RGB or grayscale images, often remain informative and can provide critical cues for navigation. For example, this is often the case when it comes to interacting with low-texture or reflective surfaces, as shown in \Cref{fig:title_image}. By learning joint representations that capture cross-modal correspondences, a navigation policy can maintain robust performance even when individual sensors fail.

The main contributions of this work are as follows.
% First, we introduce a novel~\ac{cmwae} integrated within a~\ac{rl} paradigm for robust autonomous navigation in environments that produce invalid or unreliable depth measurements.
First, we introduce a novel~\ac{cmwae}. Our approach explicitly learns cross-modal correspondences between grayscale and depth modalities by encoding them into a shared latent space, enabling the system to compensate for degraded depth information through grayscale images. 
% Even for robots with RGB-Depth sensors, we convert RGB to grayscale as we focus on such data remaining informative where depth is degraded, while opting for reduced computational complexity. 
Even for a robot equipped with an RGB-D sensor, the RGB input is converted to grayscale to preserve informative visual cues under depth degradation while reducing computational complexity.
A cross-modal training strategy is employed such that the encoder learns to extrapolate depth from features in grayscale, while a Wasserstein distance objective is used to improve generalization and consistency of the latent space.
Second, by integrating this cross-modal latent representation with the current robot state, we train a~\ac{rl}-based navigation policy using~\ac{ppo} that keeps robustness despite spatial gaps or corruption in depth measurements and allows the agent to navigate safely.
We demonstrate the method's performance through both simulation studies and experimental deployment on a flying robot.
 %This work represents, to our knowledge, the first cross-modal navigation policy that addresses depth measurement degradation.
 % \textcolor{red}{To the best of authors' knowledge, this constitutes the first cross-modal navigation policy for robust navigation under degraded depth measurements.}

The remainder of this paper is organized as follows. Section~\ref{sec:relatedwork} outlines related work, followed by the problem formulation in Section~\ref{sec:problem}. The proposed approach is presented in Section~\ref{sec:approach}. Results are detailed in Section~\ref{sec:results}, before conclusions are drawn in Section~\ref{sec:conclusions}.

\section{RELATED WORK}\label{sec:relatedwork}

Learning-based approaches that integrate perception and control have become increasingly central to autonomous navigation in complex environments. End-to-end learning approaches have emerged as alternatives to classical navigation pipelines by directly mapping sensory observations to control commands. DroNet \cite{loquercio2018dronet} pioneered this direction by employing a lightweight residual CNN to predict steering angles and collision probabilities from monocular images, enabling real-time flight on resource-constrained platforms. Building on this, \cite{gandhi2017learningflycrashing} demonstrated that collision avoidance policies could be learned through direct interaction with the environment by collecting monocular images from crashes.
% While these methods bypass explicit environment modeling, they typically rely on single sensory modalities and lack mechanisms to handle sensor failures.
To address generalization across diverse scenarios, \ac{rl}-based approaches have shown promise in learning robust navigation policies. For instance, \cite{song2023learningperceptionawareagileflight} combined visual representation learning with \ac{rl} to achieve agile flight behaviors beyond traditional control approaches. Similarly, \cite{kulkarni2024rl} proposed a modular framework using \ac{vae} to compress depth observations into collision-relevant latent features for \ac{rl}-based navigation. More recently, privileged learning frameworks \cite{wang2024visionbaseddeepreinforcementlearning}, where the critic receives noise-free perception during training, have been proposed to improve robustness at inference time. 
Despite these advances, these methods still process single modalities and do not explicitly model cross-modal relationships to compensate for sensor degradation. 

The limitations of single-modality approaches have motivated extensive exploration of multi-modal sensing in robotics \cite{delaune2019thermal, de2018robust}. Classical approaches typically employ sensor fusion techniques, such as Kalman filtering, or probabilistic frameworks that combine measurements from multiple sensors \cite{khattak2018vision}. More recently, learning-based methods have begun exploring neural architectures for fusing visual, depth, and inertial data \cite{shamwell2019unsupervised}. However, sensor degradation remains a critical challenge for autonomous systems operating in real-world environments \cite{zhang2023perception, matos2024survey}. Existing approaches address robustness primarily through data augmentation during training \cite{wang2025enhancing} or uncertainty quantification\cite{Michelmore2019UncertaintyQW}, but these strategies offer limited protection when sensors fail unexpectedly.
Cross-modal learning offers a promising approach by exploiting relationships across sensory modalities to enhance perception. In computer vision, this paradigm has led to methods for RGB-to-depth prediction \cite{ma2018sparsetodensedepthpredictionsparse}, depth completion \cite{hou2022learningefficientmultimodaldepth}, and cross-modal representation learning \cite{wang2025cross}. These approaches often align complementary information across modalities to enhance vision-based tasks \cite{gupta2016cross, huang2021multi, hoffman2016cross}. 
% Wasserstein autoencoders \cite{tolstikhin2018wasserstein}, which learn robust latent representations by minimizing optimal transport distances, have shown particular promise in this context. 
Wasserstein autoencoders \cite{tolstikhin2018wasserstein} are known to yield higher quality and more stable latent representations than variational autoencoders, owing to their optimal-transport–based objective. However, despite these advantages, their use in robotics and navigation settings remains limited, with most applications focusing on computer-vision tasks \cite{10376729, xu2019stackedwassersteinautoencoder}.
Despite these advances in computer vision, cross-modal learning has received limited attention in robotic navigation, particularly for handling sensor degradation. Our work bridges this gap by integrating \ac{cmwae} with \ac{rl} to achieve navigation robustness under degraded depth sensing, enabling policies that can compensate for depth sensor failures through alternative visual modalities.

\section{PROBLEM FORMULATION}\label{sec:problem}

This work addresses the problem of autonomous collision-avoidant navigation in unknown environments with degraded depth measurements. The robot is equipped with onboard sensors providing depth and grayscale images, along with state estimates. 
We denote the inertial frame as $\mathcal{I}$, the flying robot body frame as $\mathcal{B}$, and the vehicle frame $\mathcal{V}$. The vehicle frame is yaw-aligned with the body-fixed frame $\mathcal{B}$ and has its $x-y$ plane parallel to the inertial frame~$\mathcal{I}$. At time $t$, the robot receives a depth image $\mathbf{I}_{d, t}$ and grayscale image $\mathbf{I}_{g, t}$, and has access to its state $\mathbf{s}_t = [\mathbf{p}_t, \mathbf{v}_t, \mathbf{q}_t, \boldsymbol{\omega}_t]$, where $\mathbf{p}_t =[p_{x,t}, p_{y,t}, p_{z,t}]^\top$ is the position in $\mathcal{I}$, $\mathbf{v}_t = [v_{x,t}, v_{y,t}, v_{z,t}]^\top$ is the linear velocity in $\mathcal{I}$, $\mathbf{q}_t$ is the attitude quaternion representing the rotation from $\mathcal{I}$ to $\mathcal{B}$, and $\boldsymbol{\omega}_t = [\omega_{x,t}, \omega_{y,t}, \omega_{z,t}]^\top$ is the angular velocity in $\mathcal{B}$. Given a 3D goal position $\mathbf{p}_{\text{goal}}$ expressed in $\mathcal{I}$, current state $\mathbf{s}_t$, and sensor input pair $(\mathbf{I}_{d, t}, \mathbf{I}_{g, t})$, the objective is to compute a control policy that generates velocity command $\mathbf{u}_t = [v_{x,t}, v_{y,t}, v_{z,t}, \omega_{z,t}]^\top$, consisting of forward and vertical linear velocities and yaw rate, all expressed in $\mathcal{B}$, such that the robot navigates to $\mathbf{p}_{\text{goal}}$ while avoiding collisions with obstacles in the environment and the grayscale image compensates for regions where depth measurements are degraded or unavailable.

\section{METHODOLOGY}\label{sec:approach}

Our framework consists of two main components: (i)~a~\ac{cmwae} for cross-modal feature extraction and (ii)~a~\ac{rl} navigation policy. The~\ac{cmwae} encodes depth and grayscale images into a unified latent space that captures complementary information from both modalities, enabling the policy to maintain performance when depth measurements are degraded. We describe each component and its integration in the following subsections. We formulate the collision-free navigation as a reinforcement learning task. The state space $\mathcal{S}$ is defined as the set of all possible agent and environment states with $\mathfrak{s}_t \in \mathcal{S}$ at discrete time $t$, the action space $\mathcal{A}$ with $\mathbf{a}_t\in\mathcal{A}$, and the observation space is $\mathcal{O}$ with each agent-received observation denoted as $\mathfrak{o}_t \in \mathcal{O}$. 
Finally, $\mathfrak{r_t} = \mathcal{R}(\mathfrak{s}_t, \mathfrak{a}_t, \mathfrak{s}_{t+1})$ where $\mathcal{R}$ is the reward function while $\mathfrak{r_t}$ is the reward collected by the agent at the time $t$.
% Finally, $\mathfrak{r}_t \in \mathcal{R}$ represents the reward function.
% Our framework consists of two primary components: a Cross-Modal Wasserstein Autoencoder (CMWAE) and a PPO-based reinforcement learning agent.

% \subsection{Cross-Modal Latent Encoding for Navigation}
\subsection{Cross-Modal Wasserstein Autoencoder}

\begin{figure*}[t]
  \centering
  \includegraphics[width=\textwidth]{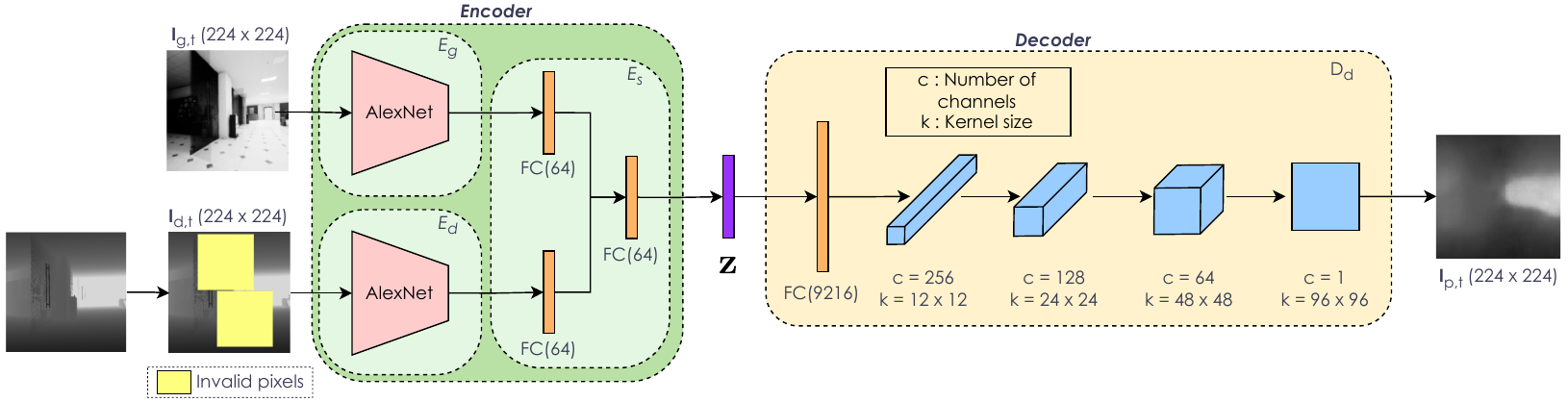}
  \caption{Architecture of the Cross-Modal Weighted Autoencoder for joint depth–grayscale representation learning.}
  \vspace{-3ex}
  \label{cmwae}
\end{figure*}

At the core of the proposed navigation framework is a \ac{cmwae} (shown in~\Cref{cmwae}), which learns a shared low-dimensional latent representation from paired grayscale and depth images. This latent space captures geometric and structural information essential for robust and collision-free navigation. The model takes as input synchronized grayscale and depth images, two inherently correlated yet complementary modalities: intensity images convey appearance and texture, while depth maps encode scene geometry.
The \ac{cmwae} employs separate encoder branches, \( E_{\text{g}} \) and \( E_{\text{d}} \), for the grayscale and depth inputs, respectively. These encoders extract low-dimensional embeddings that are passed through a shared projection network \( E_{\text{s}} \). It consists of individual \ac{fc} layers for each embedding and then a shared \ac{fc} layer which produces the fused latent vector $\mathbf{z} \in \mathbb{R}^{L_{\text{dim}}}$, where $L_{\text{dim}}$ is the size of the latent vector. 
% A latent dimension of $64$ provides a practical balance between representational capacity and computational efficiency and has been shown to be sufficient for navigation tasks in prior work~\cite{kulkarni2024rl}.
This shared latent space is passed to decoder \( D_{\text{d}} \) to reconstruct the depth modality, ensuring that the latent representation retains sufficient geometric information.
Both encoder branches $E_{\text{g}}$ and $E_{\text{d}}$ adopt the AlexNet convolutional architecture~\cite{NIPS2012_c399862d}, chosen for its proven effectiveness in visual feature extraction and its computational efficiency. Following the original AlexNet preprocessing protocol, grayscale and depth images are resized to $224 \times 224$ before being fed to the encoders. 

% The resulting feature maps are flattened and projected into the latent space. 
% The shared projection $E_{\text{s}}$ is implemented as an~\ac{mlp} that outputs the fused latent vector $\mathbf{z} \in \mathbb{R}^{64}$. 
The decoder $D_{\text{d}}$ first maps this latent vector $\mathbf{z}$ to a tensor of size $256 \times 6 \times 6$ using a \ac{fc} layer, followed by four convolutional layers to reconstruct the depth image.
The \ac{cmwae} is trained on indoor trajectories from the TartanAir dataset~\cite{tartanair2020}. We picked the TartanAir dataset due to its synchronized RGB–depth image pairs, and we then process the RGB image to obtain a grayscale image to form our dataset. Training minimizes a combination of reconstruction loss and latent regularization. The reconstruction loss, $\mathcal{L}_{\text{rec}}$, is defined as the expected squared error between the reconstructed and original depth image:
\begin{align}
\mathcal{L}_{\text{rec}} = \mathbb{E}_{\mathbf{z} \sim E(\mathbf{I}_{d}, \mathbf{I}_{g})} \| D_{d}(\mathbf{z}) - \mathbf{I}_d \|^2,
\end{align}
where $E$ denotes the complete encoder composed of $E_d$, $E_g$, and $E_s$ (green block in \Cref{cmwae}). Reconstruction is applied only to the depth modality, encouraging the network to implicitly extract geometric information from grayscale features.

To ensure a consistent latent representation, the latent space is regularized using the Wasserstein distance between the aggregated posterior of the encoder outputs and a unit Gaussian prior, following the \ac{wae} formulation with~\ac{mmd}~\cite{tolstikhin2018wasserstein}. This regularization enforces distributional alignment across modalities and improves generalization compared to conventional \ac{vae}-based schemes. The latent regularization loss is:
\begin{align}
\mathcal{L}_{\text{MMD}} = \text{MMD}(p(\mathbf{z}), \mathcal{N}(0, I))
\end{align}
The total loss combines both terms:
\begin{align}
\mathcal{L}_{\text{CMWAE}} = \mathcal{L}_{\text{rec}} + \lambda \mathcal{L}_{\text{MMD}},
\end{align}
where $\lambda > 0$ controls the contribution of latent regularization and $I$ is the identity matrix of size $L_{\text{dim}} \times L_{\text{dim}}$. %This training strategy produces a semantically meaningful and geometrically consistent latent representation suitable for navigation tasks.

We deliberately corrupt the depth inputs during training by introducing randomly positioned invalid patches to enhance robustness to depth sensor failures. We applied two different corruption schemes to the depth images in the dataset and trained the same network architecture separately under each corruption scheme. That is, one instance of the encoder–decoder model was trained exclusively on corruption scheme 1, and a second instance was trained exclusively on corruption scheme 2. This allows us to assess how each corruption scheme shapes the learned representations.

\paragraph{Corruption scheme 1 ($S_1$)}
Square patches cover between $25\%$ and $50\%$ of each image. Patch side lengths are sampled from a discrete uniform distribution $\mathcal{U}(50, 100)$ pixels and applied successively on the processed input to the network, until the target corruption coverage is reached. To ensure that the \ac{cmwae} remains effective under varying corruption levels, corruption is distributed across the dataset as follows: $5\%$ of images are corrupted at $25\%$, $10\%$ at $30\%$, $15\%$ at $35\%$, $20\%$ at $40\%$, $20\%$ at $45\%$, and the remaining $30\%$ at $50\%$. This distribution places a greater proportion of samples at higher corruption levels, providing more exposure to severe degradation during training.

\paragraph{Corruption scheme 2 ($S_2$)}
% Square patches with a fixed side length of $100$ pixels cover between $20\%$ and $50\%$ of each image. Half of the dataset is uniformly sampled to receive these corruptions, while the remaining half is left uncorrupted for both modalities.

Square patches with side length $100$ pixels cover $20\%$-$50\%$ of each corrupted depth image, with corruption levels uniformly sampled. Half the dataset is corrupted while the other half remains pristine for both modalities.

These corruption schemes ensure that even when depth information is partially missing, reconstruction from the latent space remains reliable, likely by indirectly inferring depth values from the corresponding grayscale input within corrupted regions. Unlike approaches that enforce full cross-modality reconstruction \cite{ngiam2011multimodal}, we apply cross-modal learning locally within the corrupted patches.

% We use the indoor trajectories from TartanAir dataset \cite{tartanair2020} to train our encoder. we balance the training data such that a fraction of the batches are presented with corrupted depth images and the remaining fraction has both modalities available without corruption. The corrupted depth has randomly placed invalid patches of size 100 x 100 pixels covering 20-50\% percent of the image.
% This strategy ensures that the CMWAE learns to produce meaningful latent representations and extrapolate depth maps from gray images in those small patches. Instead of employing complete cross-modality as demonstrated in \cite{ngiam2011multimodal}, we enforce cross-modality only in these patches within the image, such that the network learns to extrapolate depth values from the corresponding grayscale image. 

\subsection{Navigation Policy Learning}
% The navigation problem is assumed to be a Partially Observable Markov Decision Process. 
% The state $\mathfrak{s}_t \in \mathcal{S}$ represents the robot’s and target poses as well as the full status of the environment. As the state cannot be directly observed, the agent get an observation $\mathfrak{o}_t \in \mathcal{O}$:
% \begin{align}
%     \mathfrak{o}_t = \{{\mathbf{n}_t}, \mathrm{n}_t, \phi_t, \theta_t ,\mathbf{v}_t, \boldsymbol{\omega}_t,\mathbf{z}_{t}, \mathbf{a}_{t-1}\},
% \label{eq:observation}
% \end{align}
% where $\mathbf{n}_t \in \mathbb{R}^3$ is a unit vector to the target location expressed in $\mathcal{V}$ and the distance $\mathrm{n}_t \in \mathbb{R}$. Next, the robot's pitch $\phi_t$ and roll $\theta_t$ angles, expressed in vehicle frame~$\mathcal{V}$ which is yaw-aligned with the body-fixed frame, and has its $x-y$ plane parallel to the inertial frame~$\mathcal{I}$. The linear velocity~$\mathbf{v}~\in~\mathbb{R}^3$ and angular velocity $\boldsymbol{\omega} \in \mathbb{R}^3$ are expressed in $\mathcal{B}$. 
% The grayscale and depth images are compressed into latent embedding $\mathbf{z}_{t}$ produced using the \ac{cmwae}, and appended to the observation vector along with the previous actions $\mathbf{a}_{t-1} \in \mathbb{R}^3$.

The navigation problem is formulated as a \ac{rl} task, where the underlying state $\mathfrak{s}_t \in \mathcal{S}$ consists of the complete robot and target poses as well as the full environment status. Since the true state cannot be directly observed, the agent receives an observation $\mathfrak{o}_t \in \mathcal{O}$ at each timestep:
\begin{align}
\mathfrak{o}_t = \{\mathbf{n}_t, \mathrm{n}_t, \phi_t, \theta_t, \mathbf{v}_t, \boldsymbol{\omega}_t, \mathbf{z}_{t}, \mathbf{a}_{t-1}\}
\label{eq:observation}
\end{align}
where $\mathbf{n}_t \in \mathbb{R}^3$ is the unit direction vector pointing toward the target location expressed in vehicle frame $\mathcal{V}$, and $\mathrm{n}_t \in \mathbb{R}$ denotes the distance to the target.
The observation includes the robot's pitch $\phi_t$ and roll $\theta_t$ angles expressed in $\mathcal{V}$. The linear velocity $\mathbf{v}_t \in \mathbb{R}^3$ and angular velocity $\boldsymbol{\omega}_t \in \mathbb{R}^3$ are expressed in the body-fixed frame $\mathcal{B}$. Visual information from grayscale and depth images is compressed into a latent embedding $\mathbf{z}_{t} $ using the \ac{cmwae}, which is concatenated with the previous action $\mathbf{a}_{t-1} \in \mathbb{R}^3$ and states $\mathbf{s}_{t}$ of the robot, to form the complete observation vector.

The policy outputs actions for the aerial robot's motion, $\mathbf{a}_t = [a_{1,t}, a_{2,t}, a_{3,t}]$, which are parameterized to represent the (i) speed, (ii) inclination of the commanded velocity
with the $x$-axis of the robot and (iii) yaw rate. Finally, these actions are converted to
velocity and yaw rate command as:
\begin{align*}
v_{x,t} &= a_{1,t} \cos{a_{2,t}}, \hspace{10mm} v_{y,t} = 0.0, \\
v_{z,t} &= a_{1,t} \sin{a_{2,t}}, \hspace{10mm} \omega_{z,t} = a_{3,t},
\end{align*}
giving the velocity command $\mathbf{u}_t$, which is executed by the robot’s low-level velocity controller.
The parameterization of the controller commands is chosen to ensure that the commanded velocity vector lies within the \ac{fov} of the depth sensors and prevents a sideways collision with the environment. 

\paragraph{Reward Function}
For each state transition, the agent is rewarded for progressing toward the goal and penalized for taking inefficient actions or collisions. Based on \cite{kulkarni2024rl}, the reward function takes the form of: 
% The reward function follows the same structure as in \cite{kulkarni2024rl}, with the total reward $R(s_{t+1} | s_t, a_t)$ composed of:
\begin{align}
    \mathcal{R}(\mathfrak{s}_t, \mathbf{a}_t) = \sum_{i=1}^{4} \lambda_i r_i + \sum_{j=1}^{2} \eta_j p_j + p_{\text{crash}},
\end{align}
where $r_i$ represents positive rewards, $p_j$ penalties, defined as:
\small
% for excessive control effort, and $p_{\text{crash}}$ penalizes collisions. They're defined mathematically as follows -
\begin{align*}
& r_1 = r(\| \mathbf{n}_t \|_2, \nu_1), \hspace{10mm} r_2 = r(\| \mathbf{n}_t \|_2, \nu_2),\\
& r_3 = \frac{|\nu_3 - \| \mathbf{n}_t \|_2|}{\nu_3}, \hspace{9mm} r_4 = \nu_4 \big( \| \mathbf{n}_t \|_2 - \| \mathbf{n}_{t-1} \|_2 \big), \\
& p_1 = \sum_k \nu_{5,k} \big( r(a_t)_k, \nu_{6,k}) - 1 \big), \\
& p_2 = \sum_k \nu_{7,k} \big( r((a_t)_k - (a_{t-1})_k, \nu_{8,k}) - 1 \big), \\
& p_{\text{crash}} = -\nu_9, 
\end{align*}
\normalsize

\small
\noindent where $k$ is the action the function \( r(x, \nu) \) is defined as
\begin{equation}
r(x, \nu) = e^{-\frac{x^2}{\nu}}.
\end{equation}
\normalsize
The parameters \( \lambda_i, \eta_j > 0 \) and \( \nu_i, \nu_{m,k} > 0 \) are tuning coefficients. Scalars \( \nu_1 \ldots \nu_4 \) and \( \nu_9 \) control distance- and penalty-related terms, while \( \nu_5 \ldots \nu_8 \) are parameters for action regularization and smoothing.

% For policy optimization, we employ the Asynchronous Proximal Policy Optimization (APPO) algorithm from Sample Factory~\cite{petrenko2020sample} to train a deep neural network policy that guides the robot toward the goal location in a collision-free manner.

% where the exponential reward and penalty functions are defined as:
% \begin{align}
% R_{\exp}(m, \beta; v) &= m \, e^{-\beta v^{2}}, \\[4pt]
% P_{\exp}(m, \beta; v) &= m \left( e^{-\beta v^{2}} - 1 \right),
% \end{align}
% with \(m\) controlling the magnitude and \( \beta_{\ast} \) controlling the exponential decay factors that regulate the sensitivity of each reward or penalty to its corresponding variable i.e. larger values of \( \beta \) produce sharper decay, making the function more sensitive to deviations.

% The weights \( \lambda_i \) and \( \eta_j \) are scalar coefficients that balance the contributions of the individual reward and penalty components, respectively. 
% These weights are treated as hyperparameters and are tuned empirically to ensure stable learning.

% The agent's policy $\pi(o_t)$ maps its inputs to a 3D control command $\mathbf{a}_t \in \mathbb{R}^3$, representing forward velocity, inclination angle in the horizontal plane and yaw rate. The action is converted to a 3 dimensional linear velocity vector and angular yaw command, which are executed by the robot’s low level velocity controller. 
\begin{figure}[t]
  \centering
  \includegraphics[width=0.98\columnwidth]{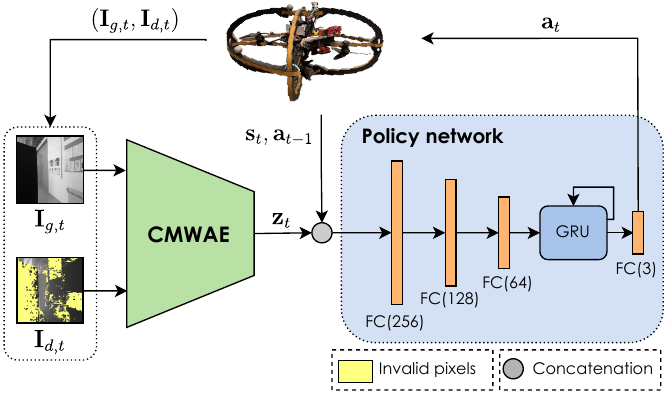}
  \vspace{-1ex}
  \caption{Depth and grayscale images are encoded into a shared latent space via the \ac{cmwae}. Later, the cross-modal representation is combined with state information and processed by the \ac{rl} policy, which outputs the velocity actions for a quadrotor platform.}
  \vspace{-3ex}
  \label{fig:RL_loop}
\end{figure}
\paragraph{Network Architecture}
We employ the \ac{ppo} algorithm from~\cite{petrenko2020sample} to train a deep neural network policy that guides the robot toward the goal location in a collision-free manner. The policy network consists of three \ac{fc} layers of size $256$, $128$, and $64$ neurons each, with an ELU activation layer, followed by a \ac{gru} to maintain temporal context. The GRU output is subsequently passed through a final \ac{fc} layer that maps it to the action vector, as illustrated in \Cref{fig:RL_loop}. Given an observation vector~$\mathfrak{o}_t$, the policy outputs a $3$-dimensional action command $\mathbf{a}_t$. For the purposes of the results presented in the next section, the speed is scaled to be between $[0, 2]$~\SI{}{\meter/\s}, inclination angle is scaled between $\pm\frac{\pi}{4}$~\SI{}{\radian}, while yaw rate is scaled between $\pm\frac{\pi}{3}$~\SI{}{\radian/\s}. 

% We use Asynchronous Proximal Policy Optimization (PPO)~\cite{schulman2017ppo} to train the navigation policy and use the Sample Factory \cite{petrenko2020sample} framework for parallelized learning. 

\paragraph{Simulation Environment}

% The reinforcement learning agent is trained within the Aerial Gym Simulator~\cite{aerialgym2023}, which provides realistic aerial robot dynamics, onboard sensors, and parallelized training. The simulation environment is modeled as a room with four walls and a floor, where obstacles are sampled from a uniform distribution $\mathcal{U}(\textcolor{red}{aaaa})$ to form cluttered navigation scenarios. To encourage generalization, the room size varies across episodes, from \(10\text{--}12\) m in length, \(5\text{--}8\) m in width, and \(4\text{--}6\) m in height. The agent is equipped with a depth camera and an RGB camera to capture exteroceptive observations in the environment. 
% Obstacles are randomly selected at with an uniform distribution from a predefined objects set consisting of planar panels\ (\(0.1 \times 1.2 \times 3.0\) m), slender cuboidal rods (\(0.1 \times 0.1 \times 2.0\) m) and thin rectangular plates (\(0.1 \times 0.5 \times 0.5\) m and \(0.1 \times 1.0 \times 1.0\) m). To maximize diversity, these obstacles are implemented as floating objects that are suspended within the environment. 

\begin{figure}[h]
    \centering
    \includegraphics[width=0.98\columnwidth]{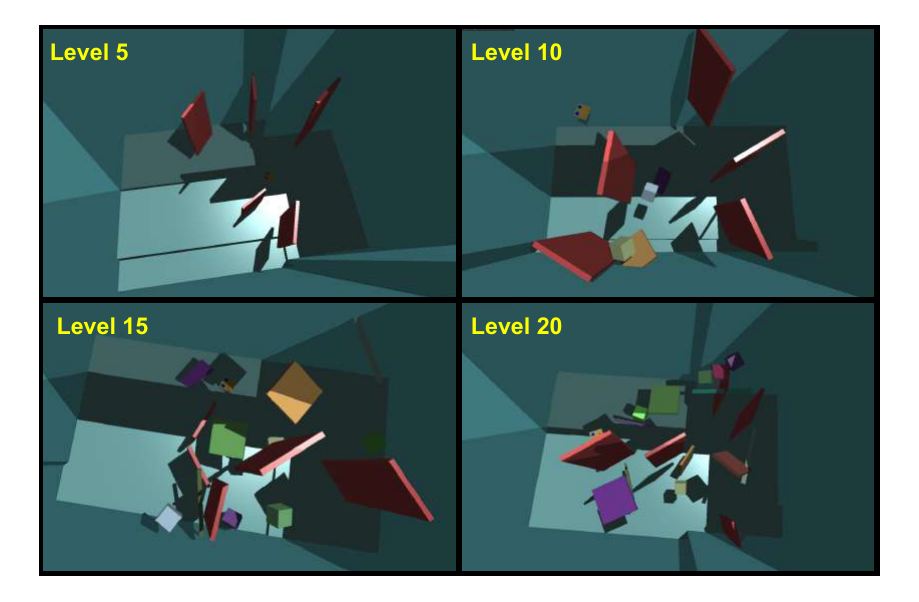}
    \caption{Top-down view of the training environment at each curriculum level.}
    \vspace{-3ex}
    \label{fig:AG_env_sample}
\end{figure}

The reinforcement learning policy is trained within the Aerial Gym Simulator~\cite{kulkarni2025aerialgymsimulatorframework}, which provides realistic aerial robot dynamics, onboard sensors, and parallelized training. 
% The simulation environment is modeled as a room with four walls and a floor, where obstacles are sampled from a uniform distribution~$\mathcal{U}(\text{predefined object set})$ to form cluttered navigation scenarios. 
The simulation environment is modeled as a room with four walls and a floor. To encourage generalization, the room dimensions are uniformly sampled at each episode reset, with length drawn from~$\mathcal{U}(10,12)$~m, width from~$\mathcal{U}(5,8)$~m, and height from~$\mathcal{U}(4,6)$~m. The agent is equipped with an image and depth sensor to capture exteroceptive observations, while attitude and velocity control are governed by a scheme relying on geometric control that builds upon the ideas in~\cite{5717652}. Obstacles are uniformly sampled from a predefined set consisting of planar panels of size $(0.1 \times 1.2 \times 3.0)$~m, slender cuboidal rods of size $(0.1 \times 0.1 \times 2.0)$~m, and thin rectangular plates of sizes $(0.1 \times 0.5 \times 0.5)$~m and $(0.1 \times 1.0 \times 1.0)$~m. To maximize environmental diversity, these obstacles are implemented as floating objects suspended within the environment. The episodes run asynchronously across environments and enable the agent to learn simultaneously across different obstacle configurations during training. 
% The environment difficulty is regulated by a curriculum schedule, where each curriculum level \(k\) introduces \(k\) obstacles into the scene. The curriculum level decreases when performance falls below a lower threshold and increases when it exceeds an upper threshold. From curriculum levels 0–5, all obstacles are planar panels; beyond level 5, obstacles are sampled from the predefined library of primitive shapes. Example environments at different curriculum levels are shown in Figure \ref{AG_env_sample}, illustrating the increase in obstacle density. During training, the network is initialized in environments at curriculum level 15, with its progression capped at level 50.
Success is defined as reaching within \SI{1}{\m} distance from the goal. Failure occurs only in the case of a collision with an obstacle. A timeout indicates that the robot remained collision-free but did not reach the goal within the episode duration. The episode length is set to $\SI{10}{\second}$, keeping in mind the dynamics of the agent and obstacle density.
Environment difficulty is governed by a curriculum schedule in which level~$k$ corresponds to $k$ obstacles. The level is decreased when the success rate across $2048$ environments falls below a lower threshold of success rate $\tau_{\mathrm{low}} = 0.30$ and increased when it exceeds an upper threshold $\tau_{\mathrm{high}} = 0.70$. Levels~$0$--$5$ use only planar panels, whereas levels greater than $5$ draw obstacles from the aforementioned predefined library of geometric primitives. \Cref{fig:AG_env_sample} illustrates the resulting increase in obstacle density with curriculum level. 

%During the start of the training, the environment is initialized at level~$15$ and its progression is capped at level~$50$.

% Once the agent achieves a success rate above 70\%, the simulator increases the curriculum level of the environment and reduces it when the crash rate exceeds 30\%. Success is defined as reaching within 1 m of the goal, while crashes are triggered by contact with any obstacle. Timeouts occur if the robot remains collision-free but does not reach the goal within the episode duration that is set to 100 timesteps, i.e, 10~\SI{}{\s}. 
% The low threshold is set to a success rate of 30\%, while the high threshold is 70\%.  Success is defined as reaching within 1~\SI{}{\m} of the goal, while crashes are triggered by contact with any obstacle. Timeouts occur if the robot remains collision-free but does not reach the goal within the episode duration set to 100 timesteps, i.e, 10~\SI{}{\s}.
% We set the lower performance threshold by $\tau_{\mathrm{low}} = 0.30$ and the upper threshold by $\tau_{\mathrm{high}} = 0.70$. 
% Success is defined as reaching within \SI{1}{\m} distance from the goal. Failure occurs only in the case of a collision with an obstacle. A timeout indicates that the robot remained collision-free but did not reach the goal within the episode duration of $\SI{10}{\second}$.
The agent begins each episode at one end of the room and is assigned a target located at the opposite end. %Both start and goal locations are sampled from a uniform distribution
% The drone's initial state and goal position are sampled from uniform distributions $\mathcal{U}(\textcolor{red}{aaaa})$ and $\mathcal{U}(\textcolor{red}{aaaa})$, respectively. 
% The drone's initial state and goal position are sampled from uniform distributions within the environment bounds. 
% In particular, the goal position is generated by drawing a ratio vector from $\mathcal{U}(\mathbf{r}_{\min}, \mathbf{r}_{\max})$, where $\mathbf{r}_{\min}$ and $\mathbf{r}_{\max}$ denote the lower and upper bounds of the normalized sampling range in $(x,y,z)$. The sampled ratio is then linearly interpolated between the environment's minimum and maximum coordinates, following AerialGym's ratio-based sampling scheme.
The start position is sampled from the first $20\%$ of the environment’s length, ensuring the drone always begins near one end of the workspace. The goal position is sampled from the last $10\%$ of the environment’s length.
To improve robustness against real-world uncertainties, we apply random forces and torques sampled from a Bernoulli process $\mathbf{B}(0.05)$ and camera pose perturbations, sampled uniformly as
$\mathcal{U}(-5, 5)$\,cm for translations and
$\mathcal{U}([-\SI{3}{\degree},\,\SI{3}{\degree}])$ for rotations. %(up to $\pm5$~cm and $\pm3^\circ$).
% The simulation also injects noise in the depth observations, sampled from a Gaussian distribution $\mathcal{N}\!\big(d,\;(\alpha d)^2\big)$, where $d$ is the clean depth and $\alpha$ is the pixel standard-deviation multiplier~\cite{khoshelham2012accuracy}. Additionally, pixels are independently dropped out with $m\sim\mathcal{B}(p_{\text{drop}})$ and set to the near–out-of-range value.

To further promote policy robustness under degraded depth sensing, a uniformly selected subset of environments receives corrupted depth inputs at each training step, while the grayscale modality remains uncorrupted throughout. The corruption is applied by overlaying randomly placed square patches with sizes uniformly sampled between \(1 \times 1\) and \(30 \times 30\) pixels. Smaller and bigger patches play a distinct role in training. Smaller patch sizes capture the fact that depth sensors often exhibit localized noise artifacts, such as speckle, missing pixels, or small dropout regions. On the other hand, larger patches reflect contiguous failures as they manifest when the sensor observes degraded areas, such as reflective surfaces or low-texture regions. The patches are applied iteratively until they collectively cover $50$\% of the depth image area.

% \begin{figure}[t]
%   \centering
%   \includegraphics[width=0.98\columnwidth]{images/ECC_cmwae_boxplots_var_compressed.pdf}
%   \caption{\ac{mse} and \ac{ssim} box plots for CMWAE trained under corruption scheme $S_1$}
%   \label{fig:boxplot_var}
% \end{figure}

% \begin{figure}[t]
%   \centering
%   \includegraphics[width=0.98\columnwidth]{images/ECC_cmwae_boxplots_fixed_compressed.pdf}
%   \caption{\ac{mse} and \ac{ssim} box plots for CMWAE trained under corruption scheme $S_2$}
%   \label{fig:boxplot_fixed}
% \end{figure}

% \begin{table}[t]
% \centering
% \caption{Average \ac{mse} ($\times 10^{-3}$) and \ac{ssim} ($\times 10^{-2}$) across corruption levels for CMWAE trained under two corruption schemes. Best values per metric are bolded.}
% \label{tab:recon-metrics}
% \renewcommand{\arraystretch}{1.15}
% \setlength{\tabcolsep}{8pt}

% \begin{tabular}{
%   c
%   c@{\,/\,}c
%   c@{\,/\,}c
% }
% \toprule
% Corruption (\%) &
% \multicolumn{2}{c}{{MSE ($\times 10^{-3}$)}} &
% \multicolumn{2}{c}{{SSIM ($\times 10^{-2}$)}} \\
% \cmidrule(lr){2-3} \cmidrule(lr){4-5}
% & $S_1$ & {\boldmath$S_2$} & $S_1$ & {\boldmath$S_2$} \\
% \midrule
% 0  & 8.49 & \textbf{5.95} & 85.29 & \textbf{85.78} \\
% 20 & 9.41 & \textbf{6.34} & 84.95 & \textbf{85.67} \\
% 30 & 10.05 & \textbf{6.72} & 84.86 & \textbf{85.57} \\
% 40 & 10.39 & \textbf{7.03} & 84.67 & \textbf{85.45} \\
% 50 & 10.71 & \textbf{8.01} & 84.79 & \textbf{85.34} \\
% \bottomrule
% \end{tabular}
% \end{table}

% \begingroup
% \color{red}

\section{EVALUATION STUDIES}\label{sec:results}

To evaluate the proposed approach, we first assess the performance for the CMWAE in cross-modal encoding before proceeding to assess the closed-loop policy results both in simulation and experimentally.

\subsection{Evaluations for CMWAE}

\begin{table}[t]
\centering
\caption{Average \ac{mse} ($\times 10^{-3}$) and Average \ac{ssim} ($\times 10^{-2}$) across corruption levels for CMWAE trained under two corruption schemes. Best values per metric are bolded.}
\label{tab:recon-metrics}
\renewcommand{\arraystretch}{1.15}
\setlength{\tabcolsep}{8pt}

\begin{tabular}{
  c
  c@{\,/\,}c
  c@{\,/\,}c
}
\toprule
Corruption (\%) &
\multicolumn{2}{c}{{MSE ($\times 10^{-3}$)}} &
\multicolumn{2}{c}{{SSIM ($\times 10^{-2}$)}} \\
\cmidrule(lr){2-3} \cmidrule(lr){4-5}
& $S_1$ & {\boldmath$S_2$} & $S_1$ & {\boldmath$S_2$} \\
\midrule
0  & 8.49 & \textbf{5.95} & 85.29 & \textbf{85.78} \\
20 & 9.41 & \textbf{6.34} & 84.95 & \textbf{85.67} \\
30 & 10.05 & \textbf{6.72} & 84.86 & \textbf{85.57} \\
40 & 10.39 & \textbf{7.03} & 84.67 & \textbf{85.45} \\
50 & 10.71 & \textbf{8.01} & 84.79 & \textbf{85.34} \\
\bottomrule
\end{tabular}
\end{table}

\begin{figure}[b]
  \centering
  \includegraphics[width=0.98\columnwidth]{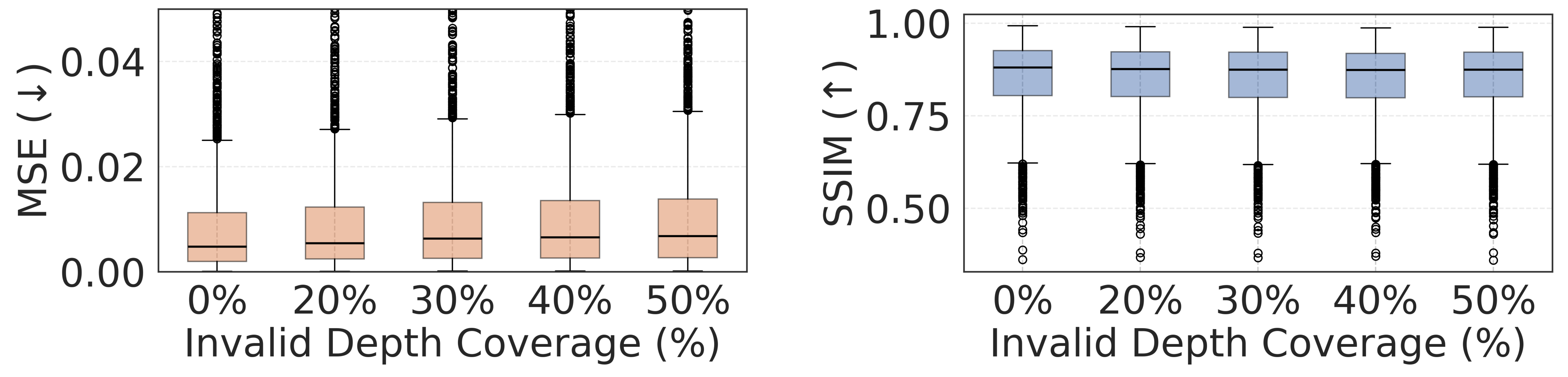}
  \caption{\ac{mse} and \ac{ssim} box plots for CMWAE trained under corruption scheme $S_1$}
  \label{fig:boxplot_var}
\end{figure}

\begin{figure}[b]
  \centering
  \includegraphics[width=0.98\columnwidth]{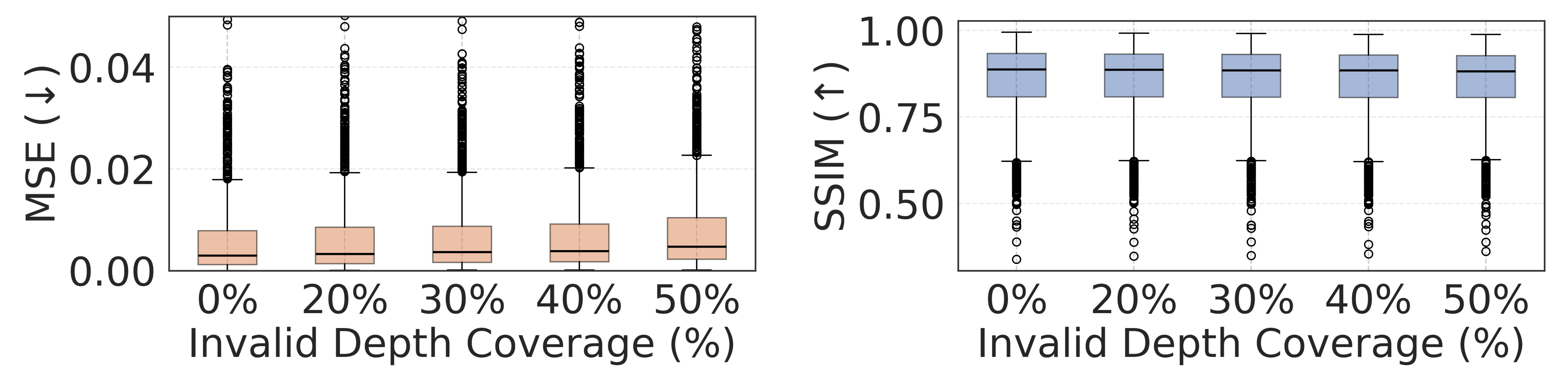}
  \caption{\ac{mse} and \ac{ssim} box plots for CMWAE trained under corruption scheme $S_2$}
  \label{fig:boxplot_fixed}
  \vspace{-2ex}
\end{figure}

For evaluation, we use a held-out test set derived from the TartanAir dataset, ensuring that none of these images are seen by the encoder during training. For all evaluations, we fix the latent dimensionality to $L_{\text{dim}} = 64$ to ensure consistency with the policy representation used later during reinforcement-learning experiments and to ensure a fair comparison with the latent-based navigation framework proposed in~\cite{kulkarni2024rl}.
We introduce corruption by overlaying randomly placed square patches (of invalid depth measurements) on each image. The patch sizes are sampled from a uniform distribution within a predefined range of $1 \times 1$ to $50 \times 50$, and patches are applied iteratively until the desired corruption coverage percentage --which ranges from $0\%$ to $50\%$-- is reached for each image. Quantitative metrics such as \ac{mse} and \ac{ssim} are then computed between the reconstructed and original images to evaluate reconstruction quality under different corruption levels. We use \ac{mse} because it provides a pixel-level assessment in the reconstructed depth map and \ac{ssim} because it offers a complementary perspective by measuring the preservation of structural information, including depth gradients and spatial relationships that are essential for accurate scene interpretation.

\begin{figure}[t]
  \centering
  \includegraphics[width=0.98\columnwidth]{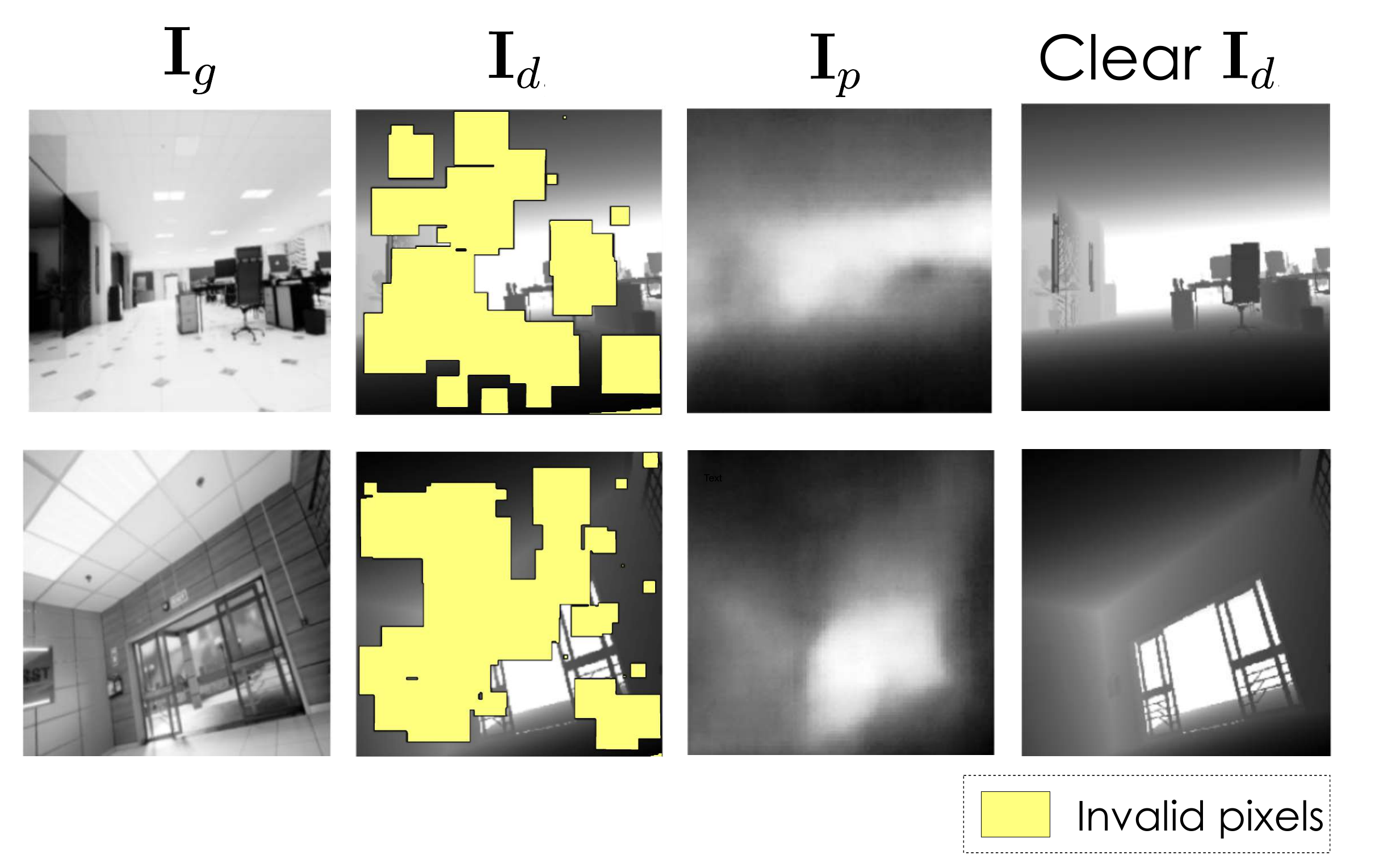}
  \caption{Example reconstructions produced by the encoder trained under corruption scheme $S_2$ in evaluation studies}
  \label{fig:recon_s2}
  \vspace{-2ex}
\end{figure}

% Table~\ref{tab:recon-metrics1} shows results for the encoder trained on corruption scheme $1$ and Table~\ref{tab:recon-metrics2} shows the corresponding results for scheme $2$.
% In addition to the tabulated results, we provide box plots of MSE and SSIM across both corruption levels in \ref{fig:mse_boxplot} and \ref{fig:ssim_boxplot} respectively, which further illustrate the distribution of reconstruction performance under each corruption scheme.
\Cref{tab:recon-metrics} presents the reconstruction results for corruption schemes $S_1$ and $S_2$. 
% To further illustrate the distribution of reconstruction performance, box plots of \ac{mse} and \ac{ssim} across both corruption schemes are provided in \Cref{fig:boxplot_var} and \Cref{fig:boxplot_fixed}.
We evaluate both models across increasing corruption levels of $0\%$, $20\%$, $30\%$, $40\%$ and, $50\%$, and the distribution of \ac{mse} and \ac{ssim} values at each level is illustrated in \Cref{fig:boxplot_var} and \Cref{fig:boxplot_fixed}. 
As the corruption level increases, \ac{mse} rises while \ac{ssim} decreases, reflecting a gradual decline in reconstruction accuracy and perceptual quality. Across all corruption levels, the model trained with scheme~$S_2$ consistently achieves lower \ac{mse} and higher \ac{ssim} than the model trained with scheme~$S_1$. Based on these findings, we select the CMWAE trained under scheme~$S_2$ for our reinforcement learning policy network. Representative examples of reconstructions generated by the encoder trained under $S_2$ are shown in \Cref{fig:recon_s2}.

\subsection{Evaluations for the RL policy in Aerial Gym}
For our evaluations, we use the Aerial Gym Simulator \cite{kulkarni2025aerialgymsimulatorframework}. 
To assess how a policy trained without accounting for sensor degradation performs under corrupted depth inputs, we compare our approach against a baseline that does not incorporate corruption during training. Specifically, depth observations remain uncorrupted throughout the training process. For this purpose, we adopt the \ac{dce}-based reinforcement learning policy (DCE-RL) \cite{kulkarni2024rl}, following the same training protocol as in the original work.
% We compare our approach against a \ac{dce}-based reinforcement learning policy \cite{kulkarni2024rl}, following the same training protocol. 
% We choose the \ac{dce}-based policy \cite{kulkarni2024rl} to serve as a baseline for comparison, since it does not account for corruption during training and allows us to evaluate the robustness of our method. 
% Specifically, depth observations remain uncorrupted during training, as in the original work. 
To ensure a fair comparison, both the DCE-RL policy and our \ac{cmwae}-based policy use the same latent size $L_{\text{dim}} = 64$.

The simulation environment is the same as described in the section \ref{sec:approach}. The policy was tested up to curriculum level 20, for sufficient complexity. We carry out the evaluations under the following conditions:
\begin{itemize}
    \item \textbf{Clean:} The agent receives depth observations directly from the simulator, without any artificial corruption throughout the episode.
    \item \textbf{Corrupted:} The agent receives depth observations that are corrupted by placing the same square patches used during policy training, randomly within the depth image and these corruptions remain for the entire episode. This aims to demonstrate performance when there are spatial gaps in depth information.
\end{itemize}

\begin{figure*}[t]
    \centering
    \includegraphics[width=\textwidth]{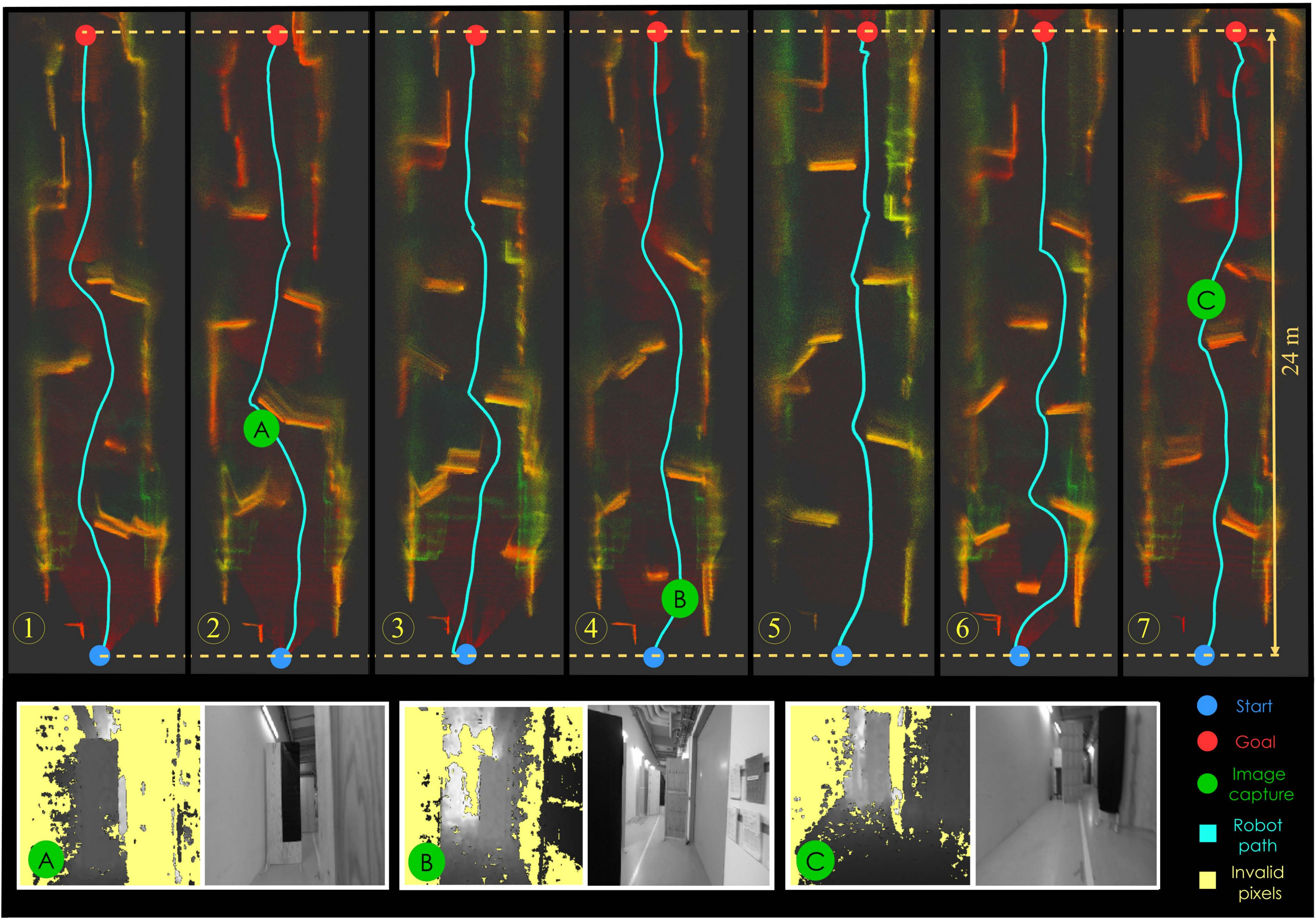}
    \vspace{-3ex}
    \caption{Top view of the trajectories of the policy across different obstacle configurations. Each column corresponds to a distinct environment with the overlaid path showing the robot’s motion from start (blue) to goal (red). The bottom row shows image captures corresponding to the points along each trajectory where the highest levels of depth corruption were encountered.}
    \label{fig:exp-trajectories}
    \vspace{-3ex}
\end{figure*}

\begin{table}[t]
\centering
\caption{RL Policy Evaluation Results. Best values per metric are bolded.}
\label{tab:rl-eval}
\renewcommand{\arraystretch}{1.15}
\setlength{\tabcolsep}{6pt}

\begin{tabular}{llccc ccc}
\toprule
 & & \multicolumn{3}{c}{DCE-RL / CMWAE-RL} \\
\cmidrule(lr){3-5}
Condition & Curr. & Success (\%) & Timeout (\%) & Crash (\%) \\
\midrule

\multirow{5}{*}{Clean}
& 0  & \textbf{99.95} / 91.05 & \textbf{0.0} / 0.43 & \textbf{0.05} / 8.53\\
& 5  & \textbf{94.00} / 85.30 & 2.88 / \textbf{1.71} & \textbf{3.19} / 13.00 \\
& 10 & \textbf{91.02} / 78.10 & 3.74 / \textbf{2.32} & \textbf{5.25} / 19.58 \\
& 15 & \textbf{85.79} / 71.51 & 5.09 / \textbf{2.50} & \textbf{9.12} / 25.99 \\
& 20 & \textbf{74.84} / 66.93 & 9.51 / \textbf{2.88} & \textbf{15.65} / 30.19\\
\midrule

\multirow{5}{*}{Corrupted}
& 0  & 0.28 / \textbf{86.02} & 72.24 / \textbf{4.64} & 27.48 / \textbf{9.34}\\
& 5  & 0.14 / \textbf{78.25} & 65.65 / \textbf{1.74} & 34.21 / \textbf{20.01}\\
& 10 & 0.04 / \textbf{68.10} & 58.27 / \textbf{1.94} & 41.69 / \textbf{29.97}\\
& 15 & 0.04 / \textbf{43.20} & 51.10 / \textbf{6.58} & \textbf{48.86} / 50.22\\
& 20 & 0.00 / \textbf{32.79} & 47.34 / \textbf{6.59} & \textbf{52.66} / 60.62\\

\bottomrule
\end{tabular}
\end{table}

\Cref{tab:rl-eval} provides a summary of the performance of both policies across the curriculum levels.
From these evaluations, we observe that under clean conditions, the \ac{dce}-based policy achieves superior performance compared to its \ac{cmwae} counterpart. The strong performance of the \ac{dce}-based policy can be attributed to the fact that \ac{dce} dedicates its entire 64-dimensional latent space to depth encoding, whereas \ac{cmwae} must allocate the same latent capacity to jointly encode both grayscale and depth information thus, effectively reducing the representational bandwidth available for each modality. 
% Additionally, \ac{dce} was partly trained directly on depth images originating from AerialGym, which gives it an inherent domain alignment that \ac{cmwae} does not possess. However, applying the same Aerial Gym data to \ac{cmwae} is not feasible, as the simulator provides textureless RGB images whose lack visual richness leads to poor reconstructions and subsequently degraded policy performance. As a result, \ac{cmwae} cannot benefit from in-domain training to the same extent as \ac{dce}.

However, when depth inputs are artificially corrupted, the \ac{dce}-based policy fails to successfully complete its objective, while the \ac{cmwae}-based policy remains robust, showing that our approach is crucial to handle degraded depth images.

\begin{table}[b]
\centering
\caption{Experimental results across 7 environments, showing time to reach the goal, path length, and average (Avg.), minimum (Min.), and maximum (Max.) corruption of the depth image.}
\label{tab:exp-results}
\renewcommand{\arraystretch}{1.1}
\setlength{\tabcolsep}{3.8pt}

\begin{tabular}{cccccc}
\toprule
\textbf{Env.} & \textbf{Time (s)} & \textbf{Path (m)} & \textbf{Avg. (\%)} & \textbf{Min. (\%)} & \textbf{Max. (\%)} \\
\midrule
1 & 20.20 & 25.05 & 29.08 & 15.34 & 49.46 \\
2 & 23.36 & 26.47 & 25.85 & 9.04  & 60.98 \\
3 & 22.00 & 25.61 & 25.04 & 12.31 & 47.76 \\
4 & 18.76 & 25.03 & 29.60 & 9.82  & 40.72 \\
5 & 21.53 & 25.75 & 26.25 & 7.64  & 50.32 \\
6 & 21.24 & 25.28 & 26.23 & 9.03  & 43.34 \\
7 & 21.17 & 25.13 & 28.72 & 14.00 & 48.46 \\
\bottomrule
\end{tabular}
\end{table}

\subsection{Real-world experiments}
% We verified the effectiveness of the method by deploying it on a quadrotor called Learning-based Micro Flyer, based on \cite{Petris2021ResilientCN} and \cite{nguyen2022motion}.
To evaluate the robustness of our approach under realistic depth sensor degradation, we conduct physical experiments using a quadrotor platform in a corridor-like environment, which is $\SI{28}{\metre}$ in length, $\SI{4}{\metre}$ in width, and $\SI{5}{\metre}$ in height. The custom-built quadrotor is equipped with an \ac{imu} and a radar sensor for state estimation ~\cite{nissov2024degradation,nissov2024robust}, and an RGB-D camera. The Intel Realsense D455 camera provides synchronized depth and color image streams at \SI{15}{\Hz}. High-level navigation runs on an onboard NVIDIA Jetson Orin NX \SI{16}{\giga\byte}, while a PX4 autopilot handles low-level control, with inter-process communication via ROS.

The experimental setup consists of $5$ to $6$ vertical panels of dimensions \SI{2.4}{\metre} $\times$ \SI{1.2}{\metre} positioned as obstacles throughout the corridor, with their configuration varied, creating different environments to test the policy's ability to generalize to unseen obstacle arrangements under depth measurement degradation.
To induce significant depth sensor degradation, we disable the infrared projector of the depth camera, which severely compromises depth measurements in regions with poor texture or ambient lighting, while the grayscale images remain unaffected, as shown in \Cref{fig:exp-trajectories}. This configuration mimics real-world scenarios where depth sensing fails due to environmental conditions, but visual information remains available. Across these environments, the depth observations contain on average $25\%$ to $30\%$ degraded pixels. \Cref{fig:RL_loop} shows the deployment loop of our policy on the platform. 
% Additionally, we also covered a panel with dark textureless fabric to worsen its observability in depth image.
To ensure that the navigation policy shall demonstrate its maneuvering capabilities for obstacle avoidance rather than simply flying over them (possible in the test space we utilized), we constrain the inclination angle of the policy to limit vertical maneuvering.
% We also constrain the maximum velocity of the policy to \SI[per-mode=symbol]{1.5}{\metre\per\second} for safety reasons. 
The start and goal locations remain consistent across all runs, with only the obstacle configuration changing, allowing us to isolate the effect of obstacle arrangement on navigation performance. The goal position is always set \SI{24}{\metre} forward and \SI{1}{\metre} upward relative to the robot's start position. 

% This setup allows us to systematically assess the policy's performance under controlled depth degradation conditions and verify that the cross-modal learning framework successfully leverages grayscale information to compensate for corrupted depth measurements and navigate in the environment. 

% For each experimental run, we collect several key metrics to assess navigation performance and safety. We record the success rate, defined as the percentage of runs where the quadrotor reaches the goal position without collisions. Additionally, we monitor the percentage of invalid depth measurements in each frame to quantify the severity of sensor degradation experienced during flight. These metrics enable comprehensive evaluation of the policy's robustness, safety, and efficiency under challenging sensory conditions.
% The degraded depth image along with the grayscale image is passed on to the \ac{cmwae} and the inferred actions are then passed on to the onboard velocity controller, provided by the autopilot. 
% \Cref{fig:RL_loop} shows the deployment loop of our policy on the platform. 
\Cref{tab:exp-results} summarizes the recorded performance across the seven environments, reporting the time required to reach the goal, the corresponding path length, and the average, minimum, and maximum percentages of corruption of the depth image observed during each run.
\Cref{fig:exp-trajectories} illustrates trajectories obtained across the seven runs. Each column corresponds to a distinct obstacle configuration, and the overlaid cyan path shows the executed motion of the robot from start (blue) to goal (red). The bottom row in \Cref{fig:exp-trajectories} shows the frames associated with the marked points along each trajectory. 
These visual examples correspond with the maximum depth corruption values summarized in \Cref{tab:exp-results} for those trajectories: $60.98\%$ in \Cref{fig:exp-trajectories}-A, 
$40.72\%$ in \Cref{fig:exp-trajectories}-B, 
and $48.46\%$ in \Cref{fig:exp-trajectories}-C.
% and generalize over environments with varying obstacle configurations. Despite the substantial degradation in the depth data, the policy is able to generate collision-free trajectories in all shown cases.
% In further detail, \Cref{tab:exp-results} summarizes the recorded performance across the seven environments, reporting the time required to reach the goal, the corresponding path length, and the average, minimum, and maximum (max) percentages of corruption of the depth image observed during each run.
% Overall, the agent consistently reaches the goal within approximately \SI{18}{\second}–\SI{23}{\second}. despite operating under varying corruption levels ranging from about $25\%$ to $30\%$. The corresponding path lengths remain tightly clustered around \SI{25}{\meter}–\SI{26}{\meter}, indicating that the agent follows similarly efficient trajectories across all runs. Notably, even in environments with higher corruption levels (e.g., $29\%$–$30\%$), the agent maintains comparable completion times and path lengths.

% Throughout the experiments, the proposed policy consistently enabled the drone to navigate successfully through each environment, without collision. 
The variation in time to reach the goal and path length across environments can be attributed to differences in obstacle configurations. Each environment presents a distinct obstacle arrangement that alters the navigable paths, leading the robot to follow slightly different trajectories.
This indicates the robust sim2real transfer and generalizability of the policy across environments with varying configurations. Despite the degraded depth images, the \ac{cmwae} encoded sufficient structural information about the obstacles to support reliable navigation performance.

\section{CONCLUSION}\label{sec:conclusions}

This paper introduced a cross-modal learning framework for autonomous navigation under depth sensor degradation. The proposed Cross-Modal Wasserstein Autoencoder enables policies to leverage grayscale information when depth measurements are corrupted, achieving robust collision-free navigation. Simulation results demonstrated improved navigation success rates and collision avoidance compared to single-modality baselines across various degradation scenarios. Experimental validation on a quadrotor platform further confirmed successful sim2real transfer and maintained performance under severe depth degradation. Future work will extend this approach to additional modalities and more complex environmental conditions.

% \endgroup


%%%%%%%%%%%%%%%%%%%%%%%%%%%%%%%%%%%%%%%%%%%%%%%%%%%%%%%%%%%%%%%%%%%%%%%%%%%%%%%%

%%%%%%%%%%%%%%%%%%%%%%%%%%%%%%%%%%%%%%%%%%%%%%%%%%%%%%%%%%%%%%%%%%%%%%%%%%%%%%%%



% \section*{ACKNOWLEDGMENT}

% The preferred spelling of the word ÒacknowledgmentÓ in America is without an ÒeÓ after the ÒgÓ. Avoid the stilted expression, ÒOne of us (R. B. G.) thanks . . .Ó  Instead, try ÒR. B. G. thanksÓ. Put sponsor acknowledgments in the unnumbered footnote on the first page.

%%%%%%%%%%%%%%%%%%%%%%%%%%%%%%%%%%%%%%%%%%%%%%%%%%%%%%%%%%%%%%%%%%%%%%%%%%%%%%%%

% References are important to the reader; therefore, each citation must be complete and correct. If at all possible, references should be commonly available publications.

% \endgroup

\bibliographystyle{IEEEtran}
\bibliography{./references}

\end{document}